\documentclass[sigconf]{acmart}

\usepackage{multirow}
\usepackage{amsfonts}

\AtBeginDocument{%
  \providecommand\BibTeX{{%
    \normalfont B\kern-0.5em{\scshape i\kern-0.25em b}\kern-0.8em\TeX}}}

\setcopyright{acmcopyright}
\copyrightyear{2021}
\acmYear{2021}
\acmDOI{10.1145/3474085.3475450}

\acmConference[MM '21]{Proceedings of the 29th ACM
International Conference on Multimedia}{October 20--24, 2021}{Virtual Event, China}
\acmBooktitle{Proceedings of the 29th ACM International Conference on Multimedia
(MM '21), October 20--24, 2021, Virtual Event, China}
\acmPrice{15.00}
\acmISBN{978-1-4503-8651-7/21/10}

\begin{document}
\fancyhead{}

\title{A Novel Patch Convolutional Neural Network for View-based 3D Model Retrieval}


\author{Zan Gao}
\affiliation{%
  \institution{Shandong Artifical Intelligence Institute, QiLu University of Technology}
  \city{Jinan}
   \country{China}
}
  \email{gaozan114@126.com}
  
\author{Yuxiang Shao}
\authornote{Corresponding Author}
\affiliation{%
  \institution{Shandong Artifical Intelligence Institute, QiLu University of Technology}
  \city{Jinan}
  \country{China}
}
  \email{shaoyuxiang93@gmail.com}
  
\author{Weili Guan*}
\affiliation{%
  \institution{School of Information Technology, Monash University}
    \city{Melbourne}
  \country{Australia}
}
  \email{honeyguan@gmail.com}

\author{Meng Liu}
\affiliation{%
  \institution{Shandong Jianzhu University} %
  \city{Jinan}
   \country{China}
}
\email{mengliu.sdu@gmail.com}

\author{Zhiyong Cheng}
\affiliation{%
  \institution{Shandong AI Institute}
    \city{Jinan}
   \country{China}
}
  \email{jason.zy.cheng@gmail.com}

\author{Shengyong Chen}
\affiliation{%
  \institution{Tianjin University of Technology}
    \city{Tianjin}
   \country{China}
}
  \email{sy@ieee.org}

\begin{abstract}
  Recently, many view-based 3D model retrieval methods have been proposed and have achieved state-of-the-art performance. Most of these methods focus on extracting more discriminative view-level features and effectively aggregating the multi-view images of a 3D model, but the latent relationship among these multi-view images is not fully explored. Thus, we tackle this problem from the perspective of exploiting the relationships between patch features to capture long-range associations among multi-view images. To capture associations among views, in this work, we propose a novel patch convolutional neural network (\textbf{PCNN}) for view-based 3D model retrieval. Specifically, we first employ a CNN to extract patch features of each view image separately. Secondly, a novel neural network module named PatchConv is designed to exploit intrinsic relationships between neighboring patches in the feature space to capture long-range associations among multi-view images. Then, an adaptive weighted view layer is further embedded into \textbf{PCNN} to automatically assign a weight to each view according to the similarity between each view feature and the view-pooling feature. Finally, a discrimination loss function is employed to extract the discriminative 3D model feature, which consists of softmax loss values generated by the fusion classifier and the specific classifier. Extensive experimental results on two public 3D  model retrieval benchmarks, namely, the ModelNet40, and ModelNet10, demonstrate that our proposed \textbf{PCNN} can outperform state-of-the-art approaches, with mAP values of 93.67\%, and 96.23\%, respectively.
\end{abstract}

\ccsdesc[500]{Information systems~Multimedia and multimodal retrieval}
\ccsdesc[500]{Information systems~Information retrieval query processing}

\keywords{3D Model Retrieval, Patch Convolutional Neural Network, Adaptive Weighted View Layer, Discrimination Loss}


\maketitle

\section{Introduction}
With the wide application of 3D systems in industrial enterprises, a huge number of 3D models have been generated and stored in enterprise repositories. As the retrieval of 3D models can save a lot of time and cost in new product development and manufacturing, it plays an important role in industrial enterprises, and  researchers have developed some 3D models retrieval methods \cite{su2015multi,feng2018gvcnn,jiang2019mlvcnn,he2019view}.
In general, the 3D model retrieval problem is formulated as follows: given a 3D model (query), the matching or relevant 3D models (documents) are retrieved, and the documents are ranked according to the similarity with the query. Since the 3D model representation plays a key important role in the 3D model retrieval, researchers often pay more attention to the 3D model representation. Then only a simple similarity metric approach, such as Euclidean distance, is utilized. The effectiveness of the 3D model representation mainly determines the success or failure of the 3D model retrieval method. Thus, with the tremendous advances of deep learning in recent years \cite{su2015multi,yang2020tree, Liu2018Cross,  feng2018gvcnn, jiang2019mlvcnn, Gao2021Pairwise, Yang2021DVMR, he2019view,Liu2019Online}, various deep networks have been employed for learning 3D model representation. According to different methods of feature representation, related methods can be roughly divided into two categories: 1) model-based methods \cite{maturana2015voxnet, qi2017pointnet++} and 2) view-based methods \cite{su2015multi, Nie2020Multigraph, feng2018gvcnn,Zhu2015Learning, jiang2019mlvcnn,he2019view, Zhou2020Semantic}. In the model-based methods, the feature is directly extracted from the 3D model, and view-based methods place multiple virtual cameras around the 3D model to generate 2D images, which are used as the input data in view-based methods. In comparison with the above model-based methods, view images can be easily obtained in the real world, and view-based methods have achieved satisfying performance in the 3D model retrieval task due to the rapid development of deep networks in 2D image analysis. Besides, the feature extraction of the 3D model is often complex, and its performance is also unsatisfactory in the 3D model retrieval task. Thus, in this work, we continue to focus on view-based 3D model retrieval.  Although many view-based 3D model retrieval approaches \cite{su2015multi,feng2018gvcnn, Liu2021Hierarchical, jiang2019mlvcnn,he2019view} have been proposed, they often focus on applying different aggregating strategies on view-level features to explore the content relationships between multiple views of a 3D model. For example, some early methods, such as MVCNN \cite{su2015multi} and MVCNN-MultiRes \cite{qi2016volumetric}, adopt view-wise pooling strategies to generate the model feature, which treat all views equally.

Nonetheless, the view-pooling scheme discards meaningful content information and spatial relationships between view images, which may help improve the performance in the retrieval task. GVCNN \cite{feng2018gvcnn} assigns different weights to views to exploit relationships among view-level features. Other methods \cite{han2018seqviews2seqlabels,dai2018siamese} utilize recurrent neural networks to obtain correlative information among view-level features. We note that these works extract the feature of each view image independently, while the latent relationships of views are left unexplored in the view feature extraction stage. Different from existing methods that aggregate view-level features, MHBN \cite{yu2018multi} employs patch-to-patch similarity measurement to obtain the model representation, but it is difficult to capture long-range associations of all multi-view images. The reason is that view features might differ significantly due to viewpoint variations. Even if there are view features with distinct differences, patches within view images might be more relevant. In comparison with utilizing view-level features directly, it is more reasonable and effective to capture long-range associations among views with patch-level features.

To solve this issue, in this work, we propose a novel patch convolutional neural network that utilizes patch-level features to learn the content information and spatial information within multi-view images. The \textbf{PCNN} takes sequential multi-view images as input, which are rendered from a circle around the 3D model. In comparison with unordered views, sequential views can help the network to learn the spatial information between adjacent views. Specifically, we first employ the CNN to separately extract patch features of each view image. Second, a novel neural network module named PatchConv is designed to capture long-range associations of all multi-view images. In detail, a k-nearest neighbor graph is constructed for each patch according to the similarity between patch features, and then, a convolution-like operation is employed to extract relational information from neighboring patches. To avoid the information loss of multiple views caused by view pooling, an adaptive weighted view layer is further embedded into the \textbf{PCNN}, which explores spatial and content relationships between adjacent view features and automatically assigns a weight to each view to make full use of the distinguishing information in the views. Finally, to improve the feature discriminability, a discrimination loss function is employed to ensure that not only model features but also view features can be distinguished, which consists of the fusion classifier and the specific classifier. The main contributions of this work are summarized as follows:

\begin{itemize}
\item We propose a novel \textbf{PCNN} method for view-based 3D model retrieval where the intrinsic relationships mining of all multi-view images, the fusion of different views, and the extraction of discriminative features are explored in a unified framework.

\item We design a patch convolutional layer to capture long-range associations among all multi-view images of a 3D model and propose an adaptive weighted view layer to automatically assign different weights to all views. Moreover, a discrimination loss function is further utilized to improve the feature discriminability of the 3D model, which consists of softmax loss values generated by the fusion classifier and the specific classifier.

\item Extensive experiments on two 3D model benchmarks show that \textbf{PCNN} can outperform state-of-the-art 3D model retrieval methods in terms of mAP.
\end{itemize}


\section{Related Work}
In the existing 3D model retrieval methods, researchers have paid more attention to the 3D model representation, and then a simple similarity metric approach such as Euclidean distance, and Cosine Distance, is used. Among the 3D model representation, since the feature extraction of 3D model features from voxels or 3D mesh or 3D point clouds \cite{maturana2015voxnet, qi2017pointnet++} is often complex, and its performance is also unsatisfied in the 3D model retrieval task, but view-based 3D model representation have achieved good performance, thus, in this work, we also focus on the view-based 3D model representation, and review it in this section.

In view-based methods, the 3D model is usually projected into a set of 2D view images. In the early development stage, hand-crafted features were often extracted to describe each view \cite{Chen2003On, Gao2020Exploring, gao20113d}, such as Zernike moments, light-field descriptors, and compact multiview descriptors. For example, Gao $et$ $al.$ \cite{gao20113d} proposed a view-based 3D model retrieval method, which employed a weighted bipartite graph matching to compare the similarity between two 3D models. Ansary $et$ $al.$ \cite{Ansary2007A} utilized adaptive views clustering, which is based on an adaptive clustering algorithm, to select the number of views with statistical model distribution scores. To remove the limitation of camera array restriction, Gao $et$ $al.$ \cite{Gao2012Camera} proposed a 3D model retrieval algorithm, in which views can be captured from any direction without camera constraint. With the development of deep neural networks, many deep learning-based methods \cite{su2015multi,dai2018siamese} have been proposed for 3D model retrieval. For example, Su $et$ $al.$ \cite{su2015multi} proposed a multi-view CNN (MVCNN) approach that first used a CNN to extract the feature of each view image individually and then max-pooled all view features into a global model feature. Further, a low-rank Mahalanobis metric was employed to improve retrieval performance. However, the simple aggregation strategy leads to the loss of meaningful information from different view images. To explore the discriminative information of different views, Feng $et$ $al.$ \cite{feng2018gvcnn} introduced a hierarchical view-group-shape architecture to assign different weights to different view images. Dai $et$ $al.$ \cite{dai2018siamese} proposed a Siamese CNN-BiLSTM network, which employed a BiLSTM to capture the relationships among view-level features. Jiang $et$ $al.$ \cite{jiang2019mlvcnn} utilized multiple groups of views from different loop directions to represent the 3D model and proposed a hierarchical view-loop-shape architecture to explore the intrinsic associations among views. Gao et al. \cite{Gao2020Exploring} systematically evaluated the performance of deep learning features in view-based 3D model retrieval on four popular datasets, meanwhile, these deep learning features were also compared with the hand-crafted features in view-based 3D model retrieval task. Inspired by triplet loss and center loss \cite{wen2016discriminative}, He $et$ $al.$ \cite{he2018triplet} proposed triplet-center loss for 3D model retrieval, which could improve the feature discriminability. Most of these approaches focus on extracting more discriminative view-level features and effectively aggregating the multi-view images of a 3D model, but the patch information of multi-view images is ignored. Thus, different from aggregating view-level features, Yu $et$ $al.$ \cite{yu2018multi} represented the 3D model from the perspective of patch-to-patch similarity measurement between all patch features. In detail, bilinear pooling was employed to aggregate patch features, and singular values of the fused feature were harmonized to generate more discriminative 3D model descriptors. However, we also observe that associations between neighboring patches in the feature space are ignored in the patch-to-patch matching. Thus, we focus on the capture of long-range associations among all multi-view images according to relational information of neighboring patches in this work.

\section{Patch Convolutional Neural Network}
In this section, we introduce the proposed \textbf{PCNN} framework in detail. As shown in Fig. 1, the framework of \textbf{PCNN} can be divided into four parts, namely, patch feature extraction, PatchConv, adaptive weighted view layer, and the discrimination loss function. A shared convolutional neural network is first employed to extract the patch features for each view image. Inspired by DGCNN \cite{wang2019dynamic} and MHBN \cite{yu2018multi}, PatchConv is designed to capture the long-range associations among all multi-view images with patch features. Moreover, an adaptive weighted view layer is further embedded into \textbf{PCNN} to assign different weights to different view images. Finally, the discrimination loss function is designed to improve the feature discriminability.

\begin{figure*}[t]
\begin{center}
\includegraphics[width=6in,height = 3in]{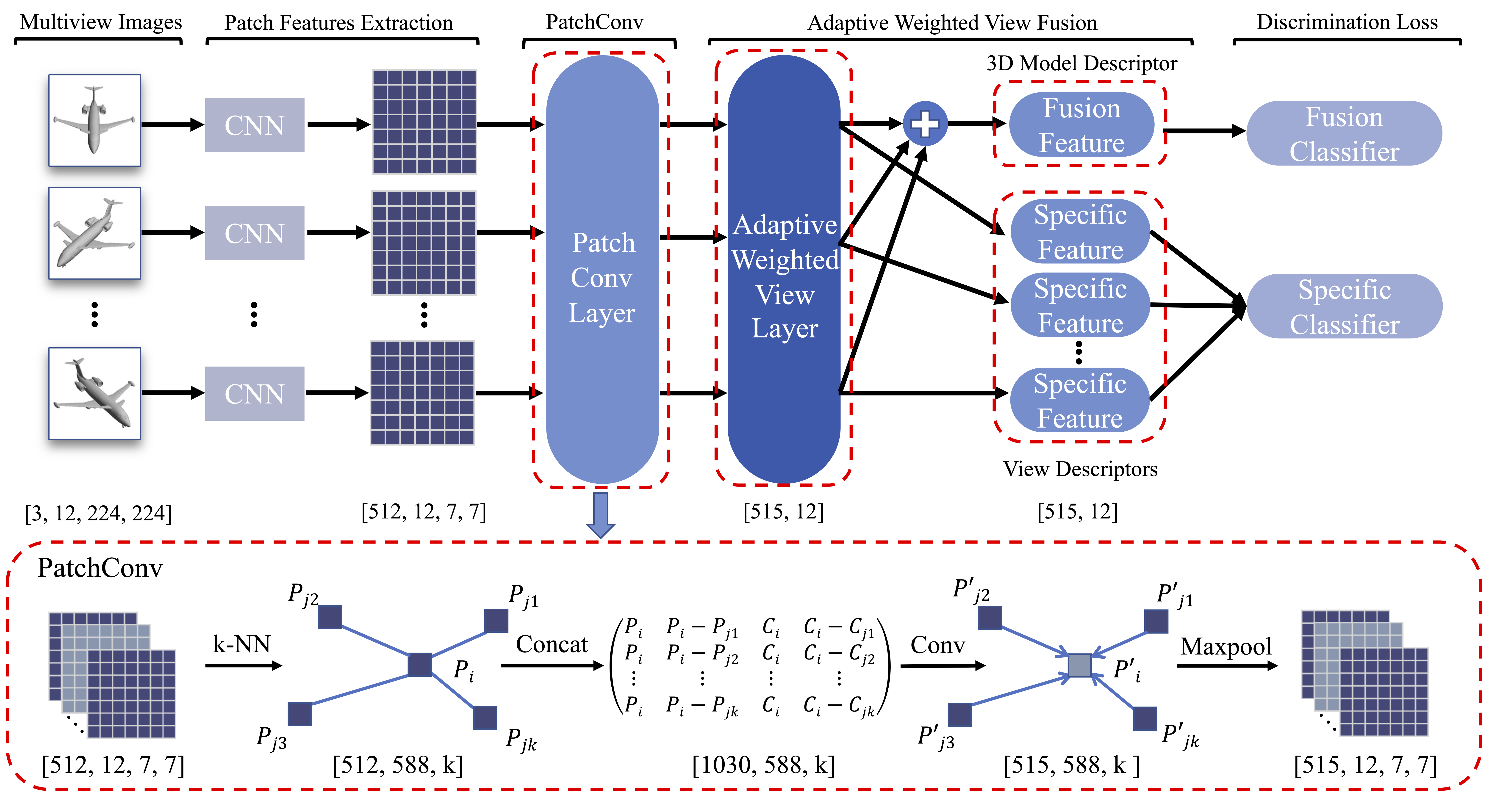}
\vspace{-1.0em}
\caption{The framework of \textbf{PCNN}}. 
\end{center}
\vspace{-1.0em}
\end{figure*}

\subsection{Patch Feature Extraction}
Given a 3D model $S$, we first generate a set of 2D greyscale images $X_S=\{x_1, x_2, ..., x_{N} \}$, where $x_i$ represents the $i^{th}$ view image and $N$ denotes the number of view images. In this work, we follow the camera setup in the MVCNN \cite{su2015multi}, which generates 12 rendered 2D view images by setting 12 virtual cameras around the model every 30 degrees. The generated view images are arranged in order. To extract patch features, ResNet34 pre-trained on the ImageNet dataset is employed as the backbone, but the last average pooling layer of ResNet34 is removed. The reason why ResNet34 is utilized as the backbone, is that overfitting often occurs due to lack of sufficient training samples, when ResNet50 or ResNet101 is employed as the backbone. Also when ResNet18 is employed, it cannot well represent each view. Thus, to balance the performance and the number of training samples, ResNet34 is utilized as the backbone. In the experiments, to match the input of ResNet34, each 2D image is first scaled to $224 \times 224$, and then these multi-view images are simultaneously fed into the backbone, thus, patch-level features can be extracted for each view. Each patch feature contains information of the corresponding region in the image. Each view image can be represented with $7 \times7$ patch features whose dimensions are $D$-dimension, and the patch features can be denoted as $p_j \in \mathbb{R}^D$. Thus, we can define $P_S = \{ p_1, p_2, ..., p_{M}\}$ as the set of patch features from all multi-view images to denote the 3D model $S$, where $p_j$ denotes the $j^{th}$ patch and $M$ represents the number of all patches of all multi-view images. Note that $7 \times7$ patches and the feature dimension of each patch depend on the structure of ResNet34. Since ResNet34 is employed, thus, in our experiments, $N = 12$, $D = 512$, and $M$ = $7 \times 7 \times N$.

\subsection{Patch Convolution Layer (PatchConv)}
In the existing view-based 3D model retrieval methods, the relationships between view-level features are fully exploited, but this is not enough. Due to viewpoint variations, view features are very different from each other, in this way, it leads to the difficulty of directly mining relationships between view-level features. Although these views are very different, there might exist more relevant patches within all views. For example, if the patch feature represents the airplane wing in a particular view, we can find other patch features that represent airplane wings in different views through the k-NN algorithm, and combining features from all views can compensate for the information loss in a particular view. Through these more relevant patches, more relational information between views can be learned. Thus, it is more reasonable to utilize patch-level features, rather than view-level features, to capture long-range associations among all multi-view images, which is the key to obtain robust 3D model features.

Inspired by DGCNN \cite{wang2019dynamic} and MHBN \cite{yu2018multi}, we propose a novel neural network module named PatchConv. Instead of directly working on view features, PatchConv exploits patch features and patch coordinates to construct multiple local neighborhood graphs in the patch feature space. The convolution-like operations are then utilized to extract relational information from the edges connecting neighboring patches in the graphs. Through this operation, we can obtain new patch-level features containing information of neighboring patches, which can make patch features more robust and capture long-range associations among all multi-view images. Supposing that there is a patch $p_i$, the $k$-nearest neighbors algorithm ($k$-NN) is employed to obtain the $k$-nearest patches among all patch features $P_S = \{ p_1, p_2, ..., p_{M}\} \in \mathbb{R}^{M \times D}$. Moreover, a graph $G(V, E)$ is further constructed, where $V = \{ 1, 2, ..., k\}$ and $E \subseteq |V| \times |V|$ denote the set of vertices and edges, respectively. In the simplest case, we construct $G$ as the k-nearest neighbors graph, and $E$ represents the edges of the form $(i, j_1), ..., (i, j_k)$ from $p_i$ to its $k$-nearest patches $p_j \in \mathcal{N}(i)$, which denotes the k-nearest neighbors of patch $i$. In PatchConv, we adopt the form $(p_i, p_j - p_i)$ to represent the features of edges connecting patch $p_i$ and its neighboring patches, which combine both the patch feature (captured by $p_i$) and local information of neighboring patches (captured by $p_j - p_i$). Due to viewpoint variations, the position of the model in different views will change, which leads to the differences between patch features in the same position of views. Considering that each patch has its particular position in the views, more information about spatial transformation, which is caused by viewpoint variation, can be obtained by utilizing patch coordinates. Thus, we further introduce $C = \{c_1, c_2, ..., c_{M}\} \in \mathbb{R}^{M \times 3}$ to represent patch coordinates, which contain spatial information of patches in the original space. Concretely, $c_i = (x, y, z)$ denotes the coordinate of the $i^{th}$ patch, and $(x, y, z)$ represents the $x^{th}$ row and the $y^{th}$ column of the $z^{th}$ view image. Thus, the new edge feature can be represented by $(p_i, p_j - p_i, c_i, c_j - c_i)$, which contains not only the information of patches in the feature space but also the coordinates of patches in the original space. The PatchConv operation can be described as
\begin{equation}
\begin{aligned}
& p'_i = {\max \limits_{j \in \mathcal{N}(i)}}{h(p_i, p_j - p_i, c_i, c_j - c_i)},
\end{aligned}
\end{equation}
where $h$ : $\mathbb{R}^{2 \times (D+3)} \to \mathbb{R}^{D + 3}$ is a parameterized nonlinear function. Here we implement function $h$ with 1 $\times$ 1 convolution, which extracts the relationships between the central patch and neighboring patches. Finally, we adopt max-pooling as the aggregation operation to generate new patch features, which can capture long-range associations among multi-view images according to the neighborhood information of patches.
\vspace{-0.5em}
\subsection{Adaptive Weighted View Layer}
Now, each view image is represented by $7 \times 7$ patch features, which contain information of neighboring patches. The average pooling layer with kernel size $7 \times 7$ is employed to generate view-level features $F_S = \{f_1, f_2, ..., f_{N}\}$, where $f_i$ represents the $i^{th}$ view feature and $N$ denotes the number of view features. Since multi-view images are sequential, features of adjacent views are relatively similar. To further mine the relationships between adjacent view images, a 1-dimensional convolution with kernel size $3 \time 3$ is utilized. In the aggregation stage of view features, max-pooling is a simple strategy to pool the view-wise features into a global model feature. The max-pooling operation only keeps the maximum activation, while the non-maximum values are ignored, which leads to sub-optimal performance. To better utilize discriminative information of different views, we design an attention mechanism to automatically assign different weights to all view images, and its pipeline is shown in Fig. 2. Concretely, the view-pooling feature $g$ is first obtained through max-pooling the view-wise features, which keeps the most prominent feature within all multi-view images. Then, we estimate the similarity score between each view feature and the view-pooling feature with cosine similarity, which is also called self-attention. The cosine similarity can be formulated as
\begin{equation}
\begin{aligned}
& s_j(f_j, g) = \frac{
\sum\nolimits_{k = 1}^D f_j(k) \times g(k)}{ \sqrt{\sum\nolimits_{k = 1}^D f_j(k)^2} \times \sqrt{\sum\nolimits_{k = 1}^D g(k)^2}},
\end{aligned}
\end{equation}
where $f_j$ represents the feature of the $j^{th}$ view image. $D$ is the feature dimension of $f_j$ and $g$, and $s_j(f_j, g)$ denotes the cosine similarity between $g$ and $f_j$. According to the similarity $s_j$, the weight $\alpha_j$ of the $j^{th}$ view is calculated as follows:
\begin{equation}
\begin{aligned}
& \alpha_j = \frac{\exp \{s_j(f_j, g ) \} }{\sum\nolimits_{i=1}^ N \exp \{s_i(f_i, g ) \} },
\end{aligned}
\end{equation}
where $N$ is the number of all multi-view 2D images. Moreover, the weights of all multi-view 2D images are further indicated by $\alpha = \{ \alpha_1, ..., \alpha_N \} $.
Finally, based on the self-attention weight of each view, we can obtain the $j^{th}$ weighted view feature $f'_j$ by
\begin{equation}
\begin{aligned}
& f'_j = \alpha_j \times f_j,
\end{aligned}
\end{equation}
where $f'_j$ is the specific view feature of the self-attention weight embedded into the view feature. The fusion feature $g'$ can be obtained by
\begin{equation}
\begin{aligned}
& g' = \sum\limits_{j = 1}^ N \alpha_j f_j,
\end{aligned}
\end{equation}
where $g'$ is the fusion feature, which is the weighted sum of the view features.

\begin{figure}[t]
\begin{center}
\includegraphics[width=3.4in,height = 1.6in]{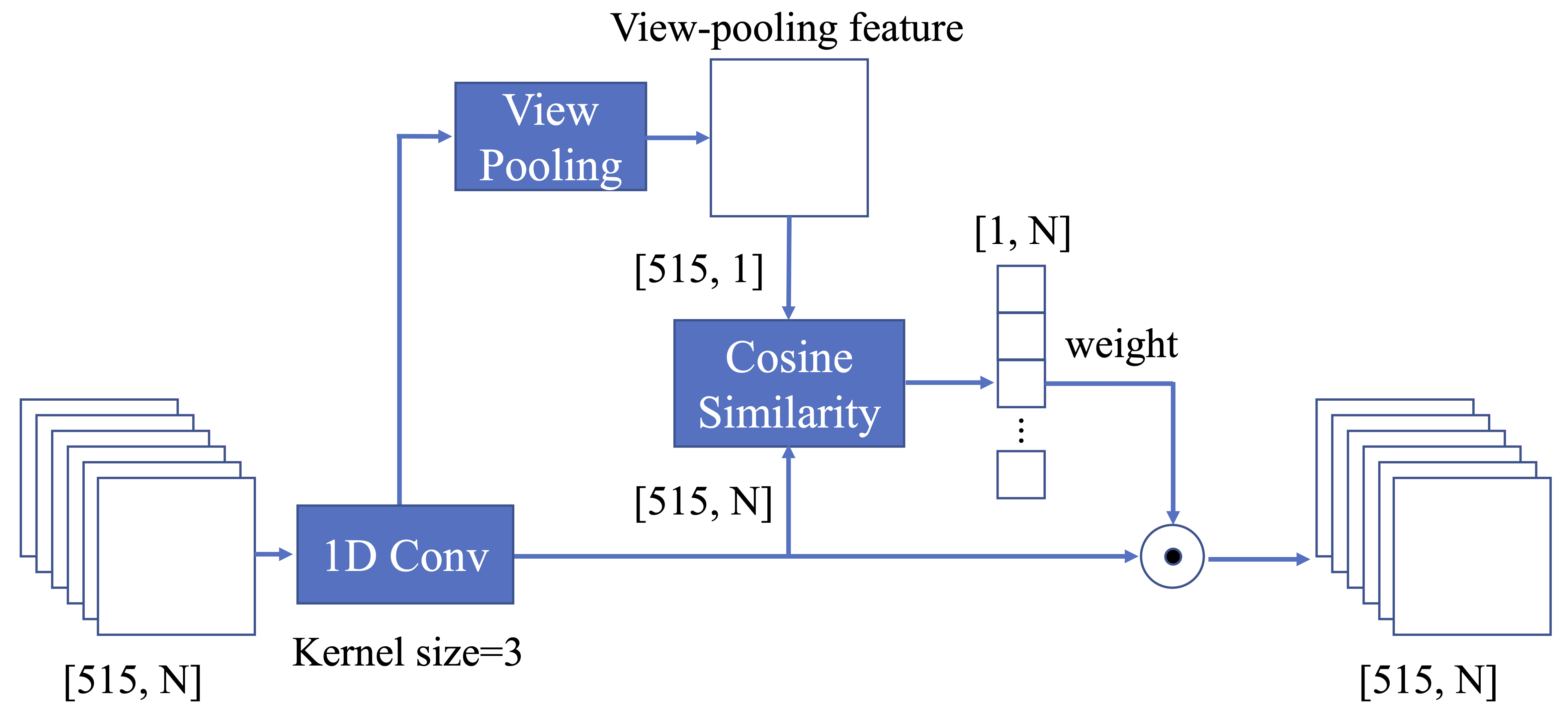}
\vspace{-1.0em}
\caption{Pipeline of adaptive weighted view layer. 1-dimensional convolution is first employed to explore relationships between adjacent views, and then, view pooling is utilized to obtain the view-pooling feature of all multi-view images. Finally, cosine similarity is used to obtain the similarity, which is the weight of each view, between the view-pooling feature and each view feature.}
\end{center}
\vspace{-1em}
\end{figure}

\subsection{Discrimination Loss Function}
In the 3D model retrieval task, the softmax loss \cite{krizhevsky2012imagenet} is usually utilized as the basic loss to increase the feature discriminability. Recently, some loss functions, such as center loss \cite{wen2016discriminative} and triplet-center loss \cite{he2018triplet}, have been employed in view-based 3D model retrieval methods to ensure that 3D model features from the same classes are pulled closer and that 3D model features from different classes are pushed away.

We note that only the 3D model feature is utilized when computing the loss values in these works, while the view features are ignored, which leads to the loss of view information. Although a single view feature cannot effectively represent the whole 3D model, it contains abundant information of the 3D model in a particular view, which can make the 3D model features easier to distinguish from other 3D models. Since each view feature can also describe the 3D model from other perspectives, we hope that these multi-view images can also be accurately recognized. To tackle this problem, a novel discrimination loss function is designed, which consists of softmax loss values generated by the fusion classifier and the specific classifier, to simultaneously classify the 3D model and each view image. To prove its effectiveness, only the simplest softmax loss is employed in the fusion classifier and specific classifier. The discrimination loss function can be defined as follows:
\begin{equation}
\begin{aligned}
& L_{Dis} = \beta \times L_{model} + \gamma \times L_{views}
\end{aligned}
\end{equation}
where $L_{Dis}$ is the discrimination loss function. $L_{model}$ indicates the loss value generated by the fusion classifier when the model feature is utilized, and $L_{views}$ denotes the loss value generated by the specific classifier when all multi-view features are used. $\beta$ and $\gamma$ are the hyperparameters used to control the contributions of $L_{model}$ and $L_{views}$.

In the calculation of $L_{views}$, the mean of losses of all multi-view features is first used, and we name it as the average view loss (AVL), which is shown as follows:
\begin{equation}
\begin{aligned}
& L_{views} = \frac{1}{N}{\sum\nolimits_{j=1}^N} L_{views}^j,
\end{aligned}
\end{equation}
where $L_{views}^j$ represents the loss value of the $j^{th}$ view image. In the experiments, we observe that some images are difficult to classify, while some images are easy to recognize. If the network can correctly recognize these difficult images, it is easier for it to classify the 3D  model. Thus, we let the network focus on recognizing these difficult images. In detail, the view weight $\alpha$ in Eq.(3) is first taken as the loss weight $\alpha'$, and then, the loss weight $\alpha'_j$ of the $j^{th}$ view image is set to $(\frac{2}{N}-\alpha'_j)$, where N denotes the number of views. In other words, when the view weight of the $j^{th}$ image is small, the image feature is very different from the view-pooling feature, and it is difficult for the specific classifier to recognize it. Therefore, we hope that the specific classifier can pay more attention to difficult views by giving their loss values larger weights. In this way, if the difficult images can be classified, the model can be easily recognized. Thus, the weighted value of $L_{views}$ can be calculated by
\begin{equation}
\begin{aligned}
& L_{views} = {\sum\nolimits_{j=1}^N} \alpha'_j L_{views}^j.
\end{aligned}
\end{equation}
We name the weighted calculation of $L_{views}$ as the weighted view loss (WVL)




\section{Experiments and Discussion}
In this section, the datasets and implementation details are first respectively introduced. The experimental results for two 3D model retrieval datasets and comparison with state-of-the-art methods are then described. Moreover, ablation studies are performed.

\subsection{Implementation Details}
To evaluate the performance of the proposed \textbf{PCNN}, we have conducted 3D model retrieval experiments on two public 3D model datasets including ModelNet40 \cite{wu20153d}, and ModelNet10 \cite{wu20153d}. ModelNet40 contains a total of 12,311 models from 40 categories,  and ModelNet10 contains a total of 4,899 models from 10 categories. In our experiments, we strictly follow the training/testing splits in MVCNN \cite{su2015multi}.  In \textbf{PCNN}, ResNet34 is utilized as the backbone. When converting the 3D model to 2D images, we strictly follow the experimental setup of MVCNN \cite{su2015multi}, and 12 virtual cameras around the model are placed every 30 degrees to generate 12 2D images. To make the network learn spatial relationships among views, the generated views are sequential. To match the input of ResNet34, each 2D image is scaled to 224 $\times$ 224. In PatchConv, $k$ of the $k$-nearest neighbors algorithm is set to 12, which means that each new patch feature aggregates the information of the 12 nearest neighbor patches. In the implementation of PatchConv, the dimension of patch features is 515 after introducing patch coordinates, and the dimension of edge features is 1,030. $1 \times 1$ convolution is employed to extract relational information from edge features, whose input dimension is 1,030, and the output dimension is 515, followed by batch normalization and LeakyReLU activation function. The parameter of LeakyReLU is set to 0.2. In the optimization of \textbf{PCNN}, Adam \cite{Kingma2014Adam} is employed as the network optimizer, where the momentum is set to 0.9. To maintain the training stability, we clip the gradients into the range [-0.01, 0.01]. The learning rate is set to $4 \times 10^{-5}$. We train the network for 30 epochs with a mini-batch size of 16. In the calculation of discrimination loss, since both the fusion classifier and specific classifier are very important, hyperparameters $\beta$ and $\gamma$ are empirically set to 0.5 and 0.5, respectively. Note that $\alpha$ doesn't need to be optimized, and it can be automatically calculated by the cosine similarity between the view-pooling feature $g$ and the feature of $j^{th}$ view image, and we also rerank the retrieval list according to classification results. Besides, in our experiments, two public evaluation metrics including Mean average precision (mAP) and precision-recall (PR) curve \cite{feng2018gvcnn, gao20113d} is utilized.

\begin{table}
\caption{Comparison with the SOTA methods on the ModelNet40 dataset}
\vspace{-1.0em}
\center
\resizebox{1.0\columnwidth}{!}{
\begin{tabular}{cccccc}
\hline
\multirow{2}{*}{Methods} & \multicolumn{2}{c} {Training Config}&\multicolumn{1}{c} {Modality}&\multicolumn{1}{c} {Retrieval}\\\cline{2-3}
&Pre-train &Fine-tune & &mAP\\
\hline
SPH\cite{kazhdan2003rotation} &$-$ &$-$ &$-$ &$33.3$\%\\
LFD\cite{Chen2003On} &$-$ &$-$ &$-$ &$40.9\%$\\
3D ShapeNets\cite{wu20153d} &ModelNet40 &ModelNet40 &Voxel &$49.2\%$\\

\hline
MVCNN\cite{su2015multi} &ImageNet1K &ModelNet40 &View &$80.2\%$\\
GIFT\cite{bai2016gift} &$-$ &ModelNet40 &View &$81.94\%$\\
GVCNN\cite{feng2018gvcnn} &ImageNet1K &ModelNet40 &View &$85.7\%$\\
\hline
SeqViews2SeqLabels\cite{han2018seqviews2seqlabels} &ImageNet1K &ModelNet40 &View &$89.09\%$\\
View N-gram\cite{he2019view} &ImageNet1K &ModelNet40 &View &$89.3\%$\\
3D2SeqViews\cite{han20193d2seqviews} &ImageNet1K &ModelNet40 &View &$90.76\%$\\
MDPCNN\cite{Gao2020Multiple} &ImageNet1K &ModelNet40 &View &$87.6\%$\\
IMVCNN\cite{He2020Improved}  &ImageNet1K &ModelNet40 &View &$90.1\%$\\
MLVCNN\cite{jiang2019mlvcnn} &ImageNet1K &ModelNet40 &View &$92.84\%$\\
\hline
\textbf{PCNN (ours)} &ImageNet1K &ModelNet40 &View &$\textbf{93.67}\%$\\
\hline
\end{tabular}}
\end{table}

\begin{table}
\caption{Comparison with the SOTA methods on the ModelNet10 dataset}
\vspace{-1.0em}
\center
\resizebox{1.0\columnwidth}{!}{
\begin{tabular}{cccccc}
\hline
\multirow{2}{*}{Methods} & \multicolumn{2}{c} {Training Config}&\multicolumn{1}{c} {Modality}&\multicolumn{1}{c} {Retrieval}\\\cline{2-3}
&Pre-train &Fine-tune & &mAP\\
\hline
SPH\cite{kazhdan2003rotation} &$-$ &$-$ &$-$ &$44.05$\%\\
LFD\cite{Chen2003On} &$-$ &$-$ &$-$ &$49.82\%$\\
3D ShapeNets\cite{wu20153d} &ModelNet10 &ModelNet10 &Voxel &$68.26\%$\\

\hline
GIFT\cite{bai2016gift} &$-$ &ModelNet10 &View &$91.12\%$\\
SeqViews2SeqLabels\cite{han2018seqviews2seqlabels} &ImageNet1K &ModelNet10 &View &$91.43\%$\\
3D2SeqViews\cite{han20193d2seqviews} &ImageNet1K &ModelNet10 &View &$92.12\%$\\
PANORAMA-ENN\cite{sfikas2018ensemble} &- &ModelNet10 &View &$93.28\%$\\
View N-gram\cite{he2019view} &ImageNet1K &ModelNet10 &View &$92.8\%$\\
IMVCNN\cite{He2020Improved} &ImageNet1K &ModelNet10 &View &$93.0\%$\\
\hline
\textbf{PCNN (ours)} &ImageNet1K &ModelNet10 &View &$\textbf{96.23}\%$\\
\hline
\end{tabular}}
\vspace{-1.em}
\end{table}

\subsection{Comparison with SOTA Methods}

In this section, we conduct a comprehensive comparison and evaluation of \textbf{PCNN}. Specifically, we first compare \textbf{PCNN} with state-of-the-art methods on two widely used datasets, the ModelNet40 and ModelNet10 datasets. Then, we present some retrieval examples for the ModelNet40 dataset. Finally, the PR curves are also given. Their results are shown in Table 1, Table 2, Fig.3, and Fig. 4, from which we make four key observations.

I) As shown in Table 1, we compare our method with competing view-based 3D retrieval methods on the ModelNet40 dataset, including SPH \cite{kazhdan2003rotation}, LFD \cite{Chen2003On}, 3D ShapeNets \cite{wu20153d}, MVCNN \cite{su2015multi}, GIFT \cite{bai2016gift}, GVCNN \cite{feng2018gvcnn}, SeqViews2SeqLabels \cite{han2018seqviews2seqlabels}, View N-gram \cite{he2019view}, 3D2SeqViews \cite{han20193d2seqviews}, MDPCNN \cite{Gao2020Multiple}, IMVCNN \cite{He2020Improved} and MLVCNN\cite{jiang2019mlvcnn}. When compared with these state-of-the-art methods, \textbf{PCNN} can obtain the best performance on both datasets. For example, in Table 1, when comparing \textbf{PCNN} with traditional methods, its performance is 60.37\%, 52.77\%, and 44.47\% higher than that of SPH, LFD, and 3D ShapeNets, respectively. Similarly, when comparing \textbf{PCNN} with deep learning methods, its performance is 13.47\%, 11.73\%, 7.97\%, 4.58\%, 2.91\%, 6.01\%, 3.57\% and 0.83\% higher than that of MVCNN, GIFT, GVCNN, SeqViews2SeqLabels, 3D2SeqViews, MDPCNN, IMVCNN and MLVCNN, respectively.

II) As shown in Table 2, we compare our method \textbf{PCNN} with competing view-based 3D retrieval methods on the ModelNet10 dataset, including SPH \cite{kazhdan2003rotation}, LFD \cite{Chen2003On}, 3D ShapeNets \cite{wu20153d}, GIFT \cite{bai2016gift}, SeqViews2SeqLabels \cite{han2018seqviews2seqlabels}, 3D2SeqViews \cite{han20193d2seqviews}, PANORAMA-ENN  \cite{sfikas2018ensemble}, View N-gram \cite{he2019view} and IMVCNN \cite{He2020Improved}. When compared with state-of-the-art methods, \textbf{PCNN} also can obtain the best performance on both datasets. For example, in Table 2, when comparing \textbf{PCNN} with traditional methods, its performance is 52.18\%, 46.41\%, and 27.97\% higher than that of SPH, LFD, and 3D ShapeNets, respectively. Similarly, when comparing \textbf{PCNN} with deep learning methods, its performance is 5.11\%, 4.80\%, 4.11\%, 2.95\%, 3.43\%, and 3.23\% higher than that of GIFT, SeqViews2SeqLabels, 3D2SeqViews, PANORAMA-ENN, View N-gram and IMVCNN, respectively.

\begin{figure}[t]
\begin{center}
\includegraphics[width=3.4in,height = 2.8in]{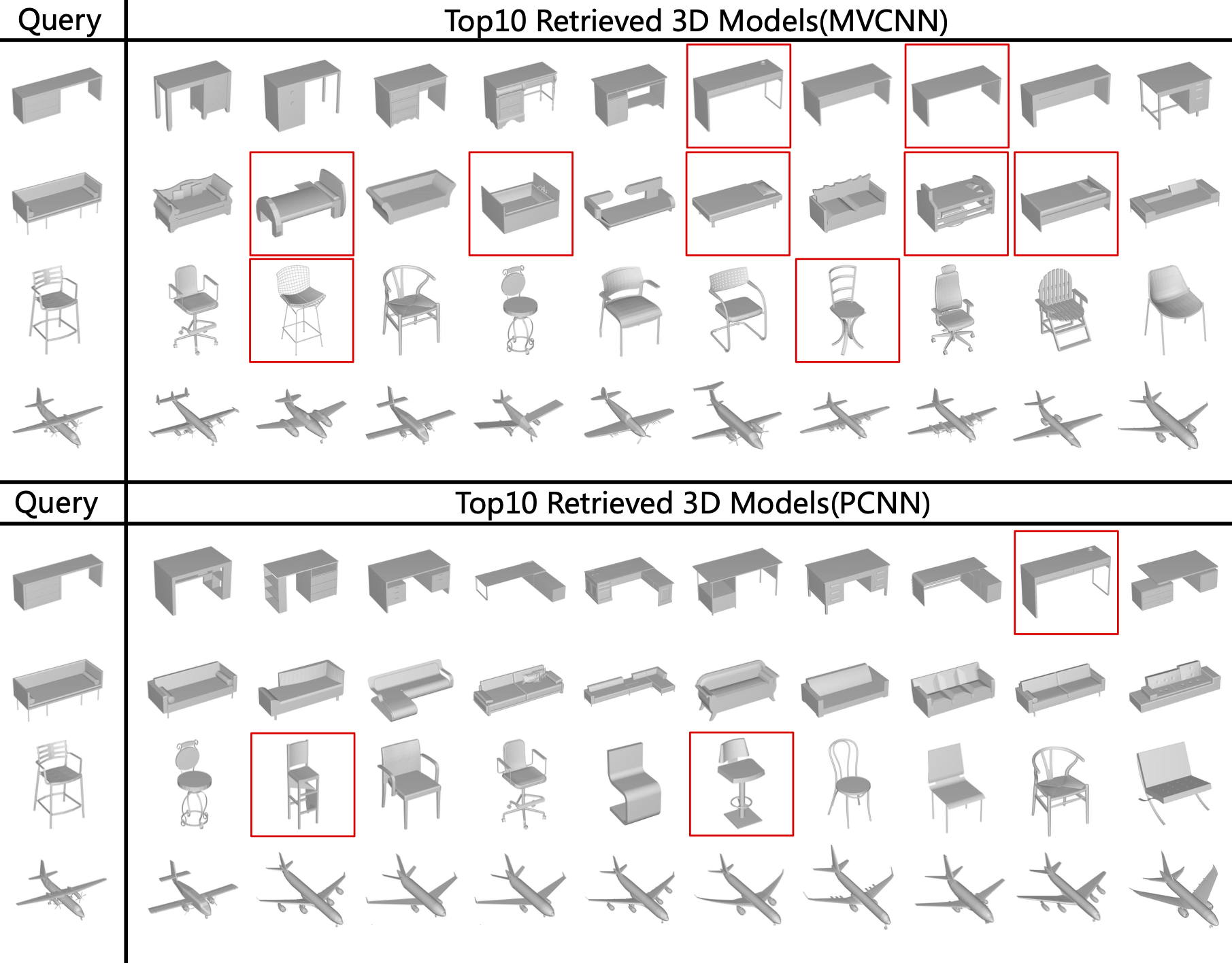}
\vspace{-1.5em}
\caption{The retrieval examples of the MVCNN and PCNN on the ModelNet40 dataset. The top 10 matching results are shown here. Note that the results of the search error are marked with red boxes. The query models are desk, sofa, chair, and airplane, respectively.}
\end{center}
\vspace{-1.5em}
\end{figure}

\begin{figure}[t]
\begin{center}
\includegraphics[width=3.2in,height = 2.3in]{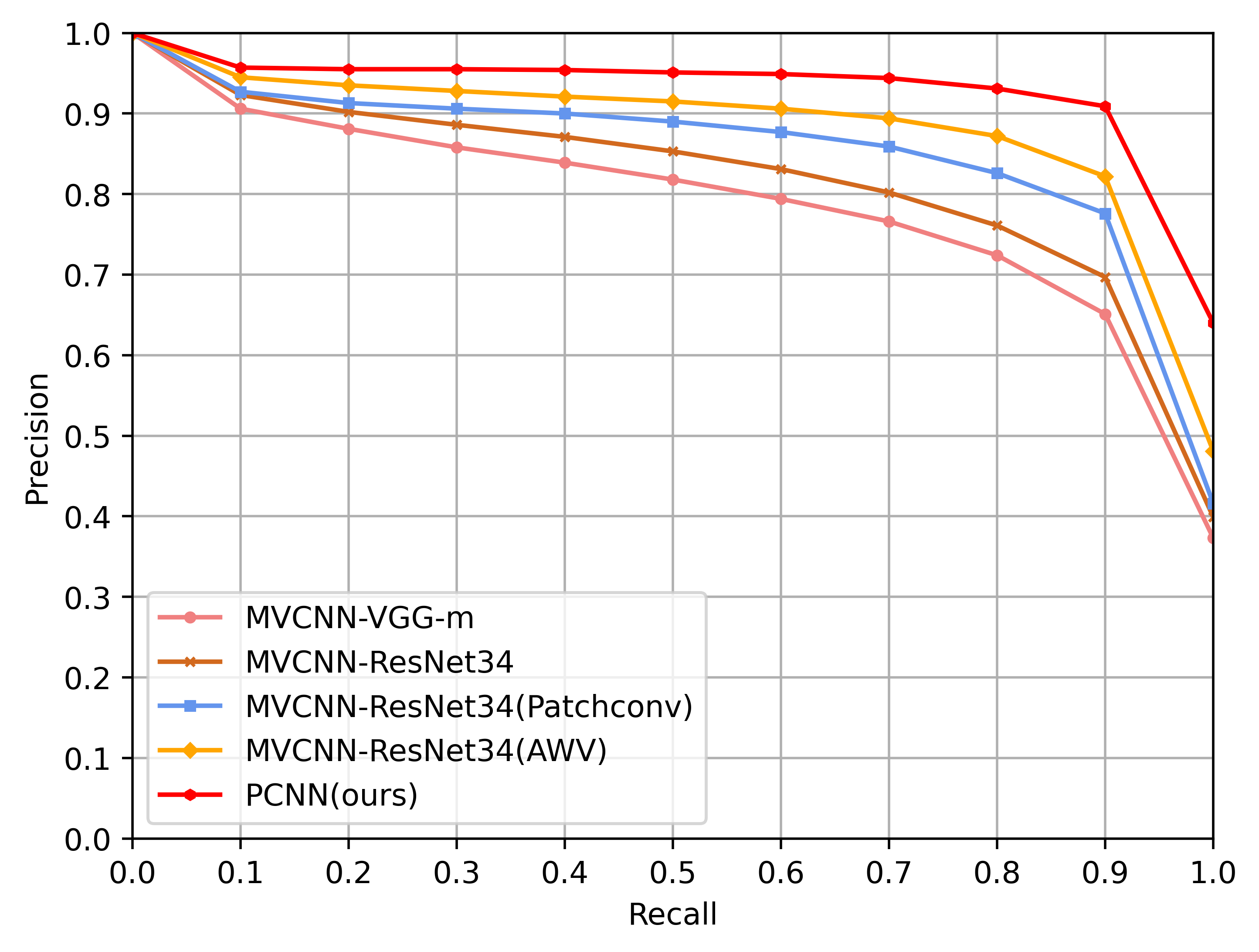}
\vspace{-1.0em}
\caption{Precision-recall curves of different modules on the ModelNet40 dataset.}
\end{center}
\vspace{-2.0em}
\end{figure}

III) In these competitors, the latent relationships among different views are explored in different ways, for example, the view-pooling operation is proposed in the MVCNN to generate the model feature, and the GVCNN distinguishes different views to mine relationships among view images. However, these issues are separately discussed, and the patch information of each view image is ignored. In our proposed \textbf{PCNN}, the patch convolution layer is designed to capture long-range associations among all multi-view images, and an adaptive weighted view layer is further embedded into the \textbf{PCNN} to automatically assign a weight to each view. Finally, a discrimination loss function is employed to extract the discriminative model feature. Most importantly, these three modules are integrated into a unified framework. Thus, the performance of \textbf{PCNN} is much better than those of MVCNN and GVCNN, and the improvements are 13.47\% and 7.97\%, respectively. Moreover, we can observe the same conclusions in Table 2 when the ModelNet10 dataset is utilized.

IV) In Fig. 3, the retrieval examples of \textbf{PCNN} and MVCNN are given, where the results of the search error are indicated with red boxes. We observe that these 3D models are very similar and that it is even difficult for people to identify them. For example, when the desk is the query (the first row in Fig.3), it is often confused with the table. Similarly, when the sofa is used as the query (the second row in Fig.3), it is often confused with the bed. Since PatchConv in \textbf{PCNN} is used to capture long-range associations among all multi-view images and a discrimination loss function is designed, the feature discriminability is improved. Thus, when \textbf{PCNN} is employed, the extracted features have very good discrimination. Although there are few retrieval errors in Top10, the number of red boxes is very small. In contrast, when MVCNN is utilized, the feature discriminability needs to be improved, and there are many retrieval errors in Top10. Thus, the proposed \textbf{PCNN} is very effective and efficient. Besides, the precision-recall (PR) curves of different modules on the MonelNet40 dataset are also given in Fig.4.

\subsection{Benefits of PatchConv and AWV}
In this section, we will assess the benifits of PatchConv and AWV. The network architecture of MVCNN is first employed, and then different backbones including VGG-m and ResNet34 are utilized in the MVCNN, and they are named MVCNN-VGG-m and MVCNN-ResNet34, respectively. To verify the validity of PatchConv and AWV, PatchConv, AWV, and the combination of both are separately embedded into the MVCNN. Finally, the performance is evaluated on the ModelNet40 dataset. The results are shown in Table 3. Note that when PatchConv, AWV, and the combination of both are embedded into MVCNN-ResNet34, they are named MVCNN-ResNet34 (PatchConv), MVCNN-ResNet34 (AWV), and PCNN (PatchConv+AWV), respectively. Note that only $L_{model}$, which is  generated by the fusion classifier, is used.

\begin{table}
\caption{Benefits of PatchConv and AWV}
\vspace{-1.0em}
\begin{center}
\resizebox{0.8\columnwidth}{!}{
\begin{tabular}{cccccc}
\hline
\multirow{1}{*}{Retrieval Methods}& &\multirow{1}{*} {mAP}\\
\hline
MVCNN-VGG-m& &80.20\%\\
MVCNN-ResNet34& &81.70\%\\
\hline
MVCNN-ResNet34 (PatchConv)&&86.28\%\\
MVCNN-ResNet34 (AWV)&&89.16\%\\
PCNN (EdgeConv+AWV)& & 92.14\%\\
\textbf{PCNN (PatchConv+AWV)}& & \textbf{92.65}\%\\
\hline
\end{tabular}}
\vspace{-2.0em}
\end{center}
\end{table}

In Table 3, we observe that MVCNN-ResNet34 can outperform MVCNN-VGG-m, whose improvement reaches 1.5\%. In other words, when ResNet34 is utilized as the backbone of  MVCNN, its retrieval accuracy is slightly better than that of VGG-m. Based on MVCNN-ResNet34, PatchConv is utilized to mine the intrinsic relationships among all different views. From Table 3, we can see that the performance of MVCNN-ResNet34 (PatchConv) is much higher than that of MVCNN-ResNet34, and the improvement achieves 4.58\%. Thus, PatchConv is very effective in mining the intrinsic relationships among different views according to patch features and patch coordinates. Also, an adaptive weighted view layer is embedded within MVCNN-ResNet34. From Table 3, we can also observe that the performance of the MVCNN-ResNet34 (AWV) is much higher than that of MVCNN-ResNet34, and the improvement reaches 7.46\%. Thus, the AWV is very efficient to automatically assign a weight to each view by the similarity between the view feature and the view-pooling feature. Besides, when both PatchConv and AWV are embedded into the MVCNN-ResNet34, the performance of PCNN (PatchConv+AWV) can be further improved.

In PatchConv, the patch coordinates are also used. In fact, this is inspired by methods based on point clouds. The point cloud is another representation of 3D models. Related methods take the coordinates of points as input and project coordinates into high-dimensional space. In the process of multi-level feature extraction, these methods combine coordinates and low-level features to generate high-level features with a multilayer perceptron. In this work, the position of the object in different images will change due to view variations, which leads to the differences between patch features in the same position. With patch coordinates, we can obtain more information about spatial transformation caused by view variation. As shown in Table 3, the performance of PCNN (EdgeConv+AWV) is 92.14\%, in which patch coordinates are not utilized. The performance of PCNN (PatchConv+AWV) is 92.65\%, which is 0.51\% higher than that of PCNN (EdgeConv+AWV).

\subsection{Effectiveness of Discrimination Loss}

To verify the validity of the discrimination loss, the model loss (ML for short) is first used in MVCNN-ResNet34 and \textbf{PCNN}, and they are marked as MVCNN-ResNet34 (ML) and PCNN (ML). The specific view loss is then added into MVCNN-ResNet34 (ML) and PCNN (ML), and their corresponding names are MVCNN-ResNet34 (ML+AVL) and PCNN (ML+AVL), where AVL indicates the average view loss in Eq.(7). Besides, when the weighted view loss (WVL) shown in Eq.(8) is utilized, we refer to it as PCNN (discrimination loss). The results are shown in Table 4. From the findings, we can observe that when AVL is used in MVCNN-ResNet34, the mAPs of MVCNN-ResNet34 (ML) and MVCNN-ResNet34 (ML+AVL) are 81.7\% and 88.39\%, respectively, and the improvement achieved is 6.69\%. Similarly, when employed in \textbf{PCNN}, the mAPs of PCNN (ML) and PCNN (ML+AVL) are 92.65\% and 93.24\%, respectively. When the discrimination loss is further used, its performance can be further improved. This result, therefore, proves the effectiveness of discrimination loss. Also, when comparing PCNN (discrimination loss) with MVCNN-ResNet34, the improvement can reach 11.97\%. This further proves the effectiveness of PatchConv, AWV and discriminative loss. These modules make the network able to capture the intrinsic relationships among different views, automatically fuse all multi-view images and extract more discriminative model features.

To further prove the effectiveness of discrimination loss, we also visualize the results of MVCNN-ResNet34, MVCNN-ResNet34 (ML+AVL), PCNN (ML), and PCNN (discrimination loss) on the ModelNet10 dataset, and their visualization results are shown in Fig. 5. The reason why we choose the ModelNet10 dataset is that the number of categories in ModelNet40 is 40, thus, it is very difficult to clearly show the results. From Fig. 5(b) and Fig. 5(d), we can observe that points from the same category are aggregative, but points from different categories are dispersed. Thus, these model features can be distinguished easily.

\begin{figure}[t]
\begin{center}
\includegraphics[width=3.4in,height = 2.4in]{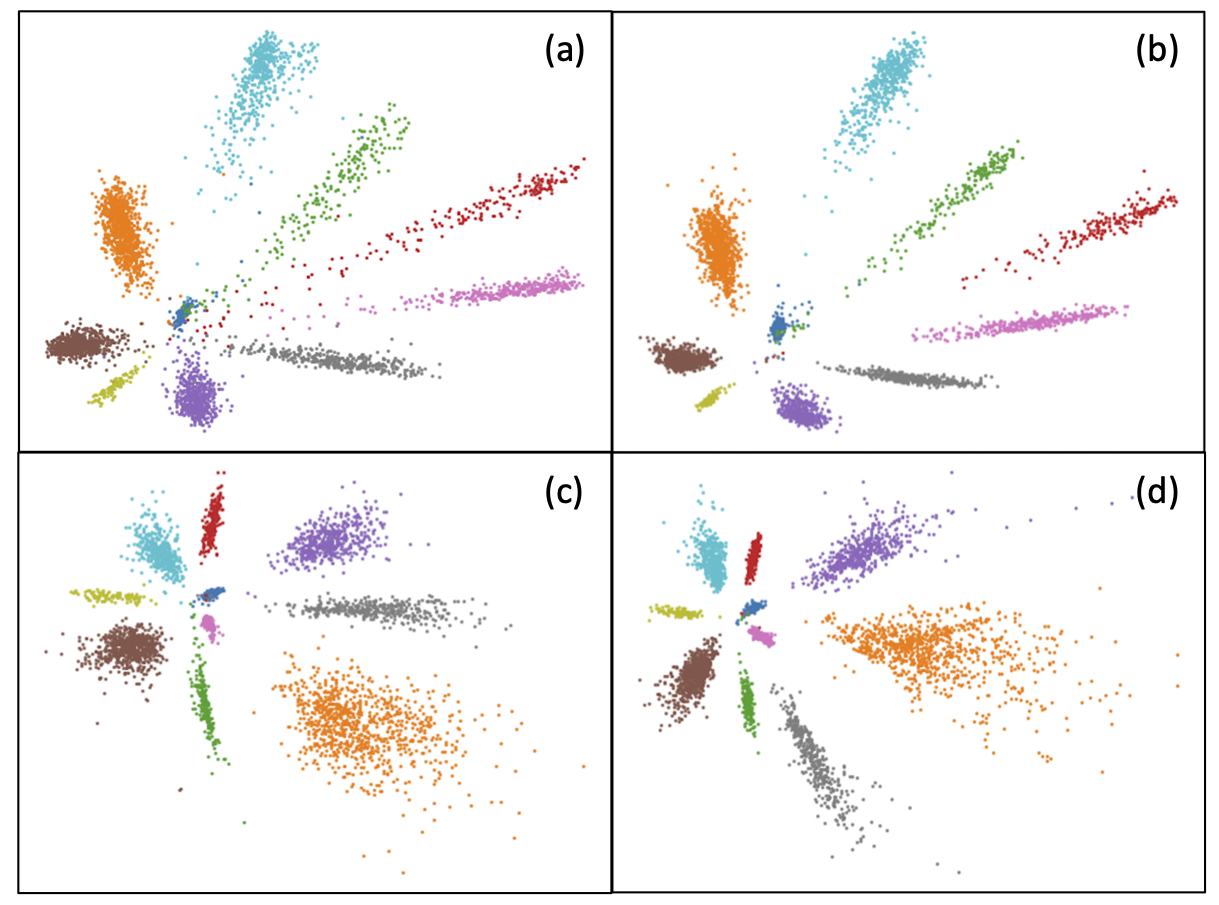}
\vspace{-2em}
\caption{Visualization results of 3D model descriptors extracted by (a) MVCNN-ResNet34 (ML), (b) MVCNN-ResNet34 (ML+AVL), (c) PCNN (ML) and (d) PCNN (discrimination loss). All training data from ModelNet10 are utilized, in which there are 10 classes in total.}
\end{center}
\vspace{-1.0em}
\end{figure}

\begin{table}
\caption{Effectiveness of Discrimination Loss}
\vspace{-1.0em}
\begin{center}
\resizebox{0.8\columnwidth}{!}{
\begin{tabular}{cccccc}
\hline
\multirow{1}{*}{Loss Function}& &\multirow{1}{*} {mAP}\\
\hline
MVCNN-ResNet34 (ML)& &81.7\%\\
MVCNN-ResNet34 (ML+AVL)& &88.39\%\\
\hline
\textbf{PCNN (ML)}& & \textbf{92.65}\%\\
\textbf{PCNN (ML+AVL)}& & \textbf{93.24}\%\\
\textbf{PCNN (discrimination loss)} & & \textbf{93.67}\%\\
\hline
\end{tabular}}
\vspace{-1.0em}
\end{center}
\end{table}

\subsection{Convergence Analysis}

\begin{figure}[t]
\begin{center}
\includegraphics[width=3.2in,height = 2.3in]{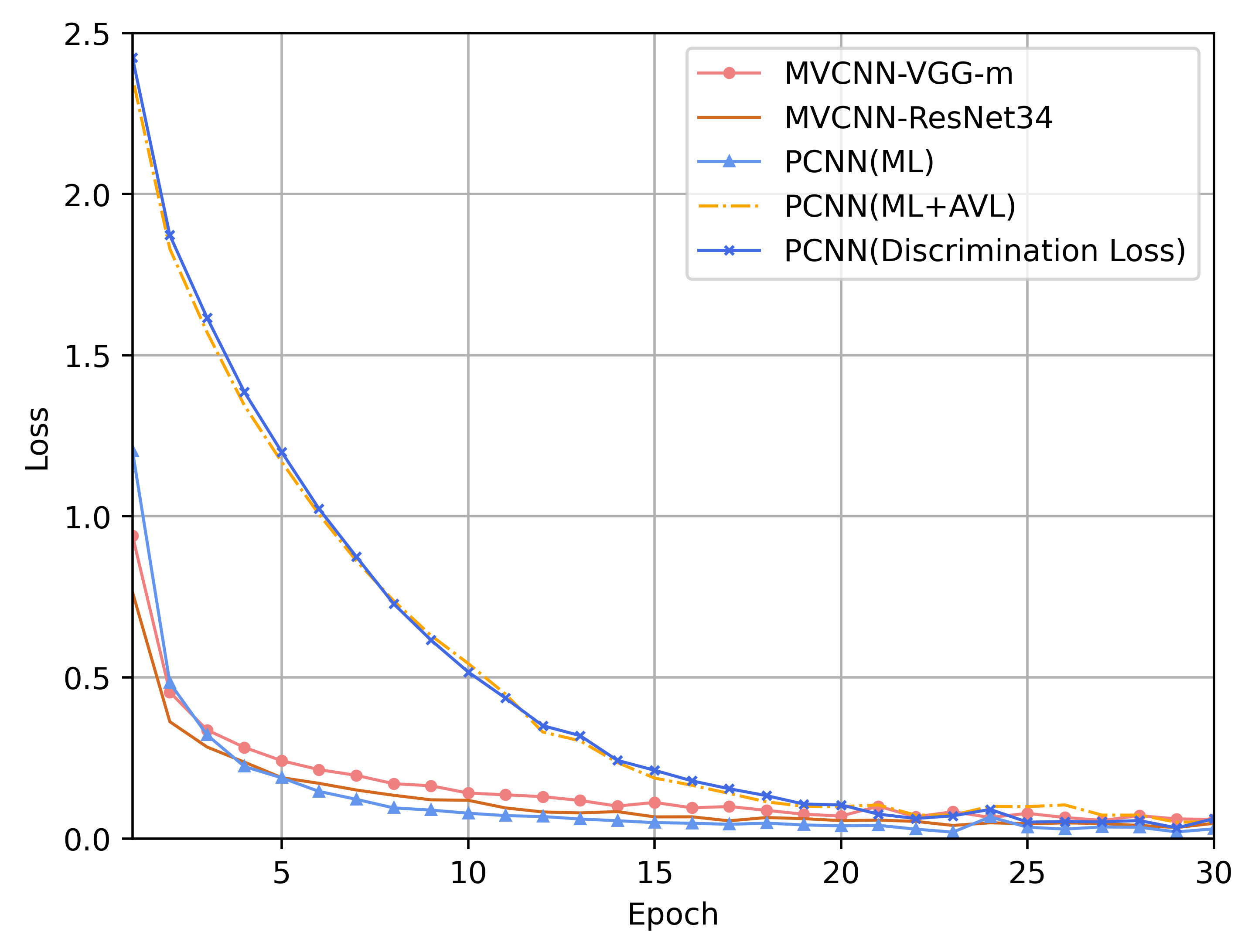}
\vspace{-1.0em}
\caption{The convergence curves of different approaches, where the horizontal axis is the number of epochs and the vertical axis is the loss value.}
\end{center}
\vspace{-1.em}
\end{figure}

In this section, we assess the convergence of the proposed \textbf{PCNN} on the ModelNet40 dataset. Moreover, we compare it with state-of-the-art approaches, and their convergence curves are shown in Fig. 6. From the results, we can see that although PatchConv and AWV are embedded into MVCNN-ResNet34, the convergence speed is also slightly quicker than those of MVCNN-ResNet34 and MVCNN. Thus, the modules we proposed are very efficient. Further, when the view loss is added into PCNN (ML), its convergence speed is slightly slower than those of PCNN (ML), MVCNN-ResNet34, and MVCNN. The reason for this result is that we classify the 3D model correctly and that all multi-view images are required to be accurately recognized in PCNN (ML+AVL) and PCNN (discrimination loss). However, PCNN (ML+AVL) and PCNN (discrimination loss) can quickly converge after only 20 epochs, which further proves the effectiveness of \textbf{PCNN}.

\section{Conclusion}
In this work, a novel \textbf{PCNN} is proposed for view-based 3D model retrieval. In \textbf{PCNN}, PatchConv and AWV are designed to exploit intrinsic relationships among all multi-view images and automatically fuse different views. Moreover, a discrimination loss function is employed to improve the feature discriminability. Extensive experiments on two public 3D model retrieval benchmarks demonstrate that \textbf{PCNN} can outperform state-of-the-art approaches. PatchConv is very effective in capturing long-range associations of all multi-view images, and AWV can automatically assign a weight to each view according to the similarity between the view feature and the view-pooling feature. Moreover, accurately classifying view features is very helpful for improving the discriminability of model features. In future work, we will focus on how to capture the relationships among all views and how to effectively fuse different views.

\section{Acknowledgement}
This work was supported in part by the National Natural Science Foundation of China (No.61872270, No.62020106004, No.92048301, No.62006142, No.61872267, No.61572357). Young creative team in universities of Shandong Province (No.2020KJN012), Jinan 20 projects in universities (No. 2020GXRC040, No.2018GXRC014).  New Artificial Intelligence project towards the integration of education and industry in Qilu University of Technology (No.2020KJC-JC01).  Shandong Provincial Key Research and Development Program (No.2019TSLH0202). 

\bibliographystyle{plain}
\balance
 \bibliography{acmmm21}

\end{document}